\documentclass[runningheads]{llncs}
\pdfoutput=1
\usepackage{times}
\usepackage{helvet}
\usepackage{courier}
\usepackage{color}
\usepackage{xspace}
\usepackage{verbatim}
\usepackage{hyperref}

\makeatletter
\newcommand{\labitem}[2]{%
\def\@itemlabel{\textbf{#1}}
\item
\def\@currentlabel{#1}\label{#2}}
\makeatother

\emergencystretch=\hsize
\usepackage[english,american]{babel}
\usepackage{graphicx}
\usepackage{mdwlist}    % for \suspend and \resume of enumerate enumeration [marcy 111909]

\usepackage{stmaryrd}      % for \inplus

\long\def\symbolfootnote[#1]#2{\begingroup%
\def\thefootnote{\fnsymbol{footnote}}\footnotetext[#1]{#2}\endgroup}

\newcommand{\bi}{\begin{itemize}}
\newcommand{\ei}{\end{itemize}}
\newcommand{\bli}{\item}

\newcommand{\tbeg}{\langle}
\newcommand{\tend}{\rangle}
\newcommand{\st}{
}

\newcommand{\suggestion}[2]{}

\def\mc{{\mathcal{C}}} 
\def\ma{{\mathcal{A}}} 

\def\no{{not} \;} 
\def\ar{\leftarrow} 
\def\beq{\begin{equation}} 
\def\eeq#1{\label{#1}\end{equation}} 
\def\beq{\begin{equation}}
\def\eeq#1{\label{#1}\end{equation}}
\def\ba{\begin{array}}
\def\ea{\end{array}}

\def\wseq{{\sc wseq}\xspace}
\def\is{{\sc is}\xspace}
\def\rf{{\sc rf}\xspace}

\def\seq{{\sc seq}++\xspace}

\def\aspcomp{{\sc aspcomp}\xspace}

\def\clasp{{\sc clasp}\xspace}

\def\bprolog{{\sc bprolog}\xspace}

\def\dlv{{\sc dlv}\xspace}

\def\cmodels{{\sc cmodels}\xspace}
\def\ezcsp{{\sc ezcsp}\xspace}
\def\clingcon{{\sc clingcon}\xspace}
\def\gecode{{\sc gecode}\xspace}
\def\acsolver{{\sc acsolver}\xspace}
\def\idp{{\sc idp}\xspace}

\def\minisat{{\sc minisat}\xspace}

\def\bbox{{\it black-box}\xspace}
\def\cbox{{\it clear-box}\xspace}
\def\gbox{{\it grey-box}\xspace}

\title{
Hybrid Automated Reasoning Tools: from Black-box to Clear-box Integration
}
\author{Marcello Balduccini\inst{1} \and Yulia Lierler\inst{2}}

\titlerunning{Hybrid Automated Reasoning Tools: from Black-box to Clear-box Integration}
\authorrunning{M.~Balduccini and Y.~ Lierler} %Marcello Balduccini \and Yulia Lierler} 
\institute{College of Information Science and Technology \\
Drexel University \\
\email{marcello.balduccini@gmail.com}
\and
Computer Science Department \\
University of Nebraska at Omaha \\
\email{ylierler@unomaha.edu}
}

\begin{document}

\selectlanguage{american}

\maketitle

\begin{abstract}
Recently, researchers in 
answer set programming and constraint programming spent
significant efforts in the development of
hybrid languages and solving algorithms combining the strengths of these
traditionally separate fields. These efforts resulted 
in a new research area: constraint answer set programming~(CASP). 
CASP languages and  systems proved to be largely successful at
providing  efficient solutions to problems
involving hybrid reasoning tasks, such as scheduling problems 
with elements of planning. Yet, the
development of CASP systems is difficult, requiring
non-trivial expertise in multiple areas.
%, as well as 
% the analysis and the comparison of competing
%approaches is a difficult task.
This suggests a need for a study identifying general
development principles of hybrid systems. Once these principles and
their implications are
well understood, the development of hybrid languages and systems
may become a well-established and well-understood routine
process.
As a step in this direction, in this paper we
conduct a case study aimed at evaluating various
integration schemas of CASP methods.
%Our experiments are conducted using a specific
%state-of-the-art solver as a testbed, and comparing
%various architectural choices.
\end{abstract}
\section{Introduction}
Knowledge representation and automated reasoning are  areas of Artificial Intelligence 
 dedicated to understanding and  automating 
 various aspects of reasoning. Such traditionally separate fields of AI
  as answer set programming~(ASP)~\cite{bre11},  propositional satisfiability~(SAT)~\cite{gom08},
 constraint (logic) programming~(CSP/CLP)~\cite{ros08,jm94} are all
 representatives of directions of research in automated reasoning.  
%These
% fields mastered the algorithms and tools to tackle the problems
% posed in the languages appropriate for each area. 
The algorithmic techniques
 developed in subfields of automated reasoning are often suitable for
 distinct reasoning tasks. For example, answer set
 programming  proved to be
 an effective  tool for formalizing elaborate planning tasks whereas
 CSP is efficient in solving difficult scheduling
 problems. Nevertheless, if the task is to solve complex scheduling
 problems  requiring elements of planning then neither ASP nor CSP 
 alone is sufficient. 
In recent years,  researchers attempted to address this problem by
developing   {\it hybrid} approaches that combine
algorithms and systems from different AI subfields. 
Research in satisfiability modulo theories~(SMT)~\cite{nie06}
is a well-known example of this trend.

\st
More recent examples include constraint answer set 
programming (CASP) \cite{lier12}, which integrates answer set 
programming  with constraint (logic) programming.
%As a declarative
%programming paradigm stemming from ASP, CASP  provides a rich, 
%simple modeling language that, among other features,
%incorporates recursive definitions.
%CASP also inherits from ASP the ability to use variables; 
%software tools called grounders are used as front
%ends of answer set solvers to eliminate variables, 
%while SAT-like procedures form their back-ends.
%From CSP/CLP languages CASP inherits the ability to
%use non-boolean constructs: not only 
%some constraints are more naturally 
%expressed by non-boolean constructs, but constraint
%programming tools often include algorithms for processing
%such constraints that outperform SAT-like solving methods.
%For instance, variables over large domains pose a challenge
%to ASP both in the grounding and in the solving stages of computation,
%whereas
%CSP/CLP provides various specialized procedures to tackle constraints
%over such variables.
 Constraint answer set programming allows to combine the best of two
  different automated reasoning worlds: 
(1) modeling capabilities of ASP together with advances of
its SAT-like technology in solving and (2) constraint processing techniques for
effective reasoning over non-boolean constructs.
This new area has 
already demonstrated promising activity, including
the development of the CASP solvers {\acsolver}~\cite{mel08},
{\clingcon}~\cite{geb09}, {\ezcsp}~\cite{bal09a}, and 
{\sc idp}~\cite{idp}. Related techniques have also been used in
the domain of hybrid planning for robotics~\cite{spe13}.
CASP opens new horizons for
 declarative programming applications. 
 Yet the developments in this field
 pose a number of questions, which also apply to the automated reasoning
 community as a whole.

\st
The broad attention received by the SMT and CASP paradigms, which aim to integrate
and build synergies between diverse constraint technologies, and the success they enjoyed
suggest a necessity of a principled and general study of methods to
develop such hybrid solvers.  \cite{lier12}
provides a study of the relationship between various CASP  
solvers  highlighting  the importance of creating unifying
approaches to describe such  systems. 
For instance,  the CASP systems \acsolver, \clingcon, and \ezcsp
came to being within two consecutive years. 
These systems  rely on different ASP and CSP technologies, so it is difficult to clearly
articulate their similarities and differences. In addition, the CASP solvers adopt  different
communication schemas among their heterogeneous solving components.
The system {\ezcsp} adopts a ``{\it \bbox}'' architecture, whereas
{\acsolver} and {\clingcon} advocate tighter integration.
   The crucial message transpiring from these developments in the CASP community is the ever
growing need for  standardized techniques to integrate computational
methods spanning multiple research areas. 
Currently such an integration  requires 
nontrivial expertise in multiple areas, for instance, in SAT, ASP and
CSP. We argue for
undertaking an effort to mitigate difficulties  of designing hybrid
reasoning systems by identifying general principles for their
development and studying the implications of various design choices. 

\st
As a step in this direction, in this paper we conduct a case study aiming
to explore a crucial aspect in building hybrid 
systems -- the integration schemas of participating solving methods.
We study various integration schemas and their performance, using CASP as our test-bed domain.  
As an exemplary subject  for our study we take the CASP system {\ezcsp}. 
\st
Originally, {\ezcsp} was developed as an inference engine for
CASP that allowed a lightweight, \bbox,
integration of ASP and CSP.  
In order for our analysis to be conclusive
we found it important to study the different integration mechanisms using the
same technology.
Within the course of this work we 
implemented  ``\gbox'' and ``\cbox'' approaches
for combining ASP and CSP reasoning  within {\ezcsp}. 
We evaluate these  configurations of {\ezcsp} on three
domains -- Weighted Sequence, Incremental Scheduling, and
Reverse Folding -- from the {\sl Model and Solve} track of 
the {\sl Third  Answer Set Programming Competition --
  2011} ({\aspcomp})~\cite{aspcomp11}. Hybrid paradigms such as CASP
allow for mixing language constructs and computational mechanisms stemming
from different formalisms. Yet, one may design encodings that favor
only single reasoning capabilities of a hybrid system. For this reason,
in our study we
evaluate different encodings for the proposed benchmarks that we call
``pure ASP'', ``true CASP'', and ``pure CSP''. As a result we expect to
draw a comprehensive picture comparing and contrasting various
integration schemas on several kinds of encodings possible within
hybrid approaches. 

\st
We start with a brief review of the CASP formalism. Then we draw a
parallel to SMT solving, aimed at showing that it is possible to transfer to SMT the results
obtained in this work for CASP solving. In Section~\ref{sec:integration} we review the
integration schemas used in the design of hybrid solvers focusing on the schemas
 implemented in {\ezcsp}  within this project.   Section~\ref{sec:domain}
provides a brief introduction to 
 the application domains considered, and discusses the variants
of the encodings we compared.
Experimental results and their analysis form Section~\ref{sec:experiments}.

\section{Review: the CASP and SMT problems}\label{sec:acprograms}
The review of logic programs with constraint atoms follows the lines of~\cite{lier-alp}.
A {\sl regular program} is a finite set of rules of the form
\beq
\ba {l}
a_0\ar a_1,\dots, a_l,not\ a_{l+1},\dots,not\ a_m,\ 
not\ not\ a_{m+1},\dots,not\ not\ a_n,
\ea
\eeq{e:rule}
where $a_0$ is $\bot$ or an atom, and
each $a_i$ ($1\leq i\leq n$) is an atom.
This  is a special case of 
programs with nested expressions \cite{lif99d}.  We refer the reader 
to~\cite{lif99d}
for details on the definition of an answer set of a logic 
program. 
A \emph{choice rule}
construct $\{a\}$~\cite{nie00} of 
the {\sc lparse} language
 can be seen as an abbreviation for a rule
\hbox{$a\ar\ \no \no a$}~\cite{fer05b}. We adopt this abbreviation.

\st
A constraint satisfaction problem (CSP) is defined as a triple
$\langle X,D,C \rangle$, where $X$ is a set of variables, $D$ is a domain of
values, and $C$ is a set of constraints. Every constraint is  a
pair $\langle t,R \rangle$, where $t$ is
an $n$-tuple of variables and $R$ is an $n$-ary relation on $D$. An
{\em evaluation}
of the variables is a function from the set of variables to the domain
of values, $\nu:X \rightarrow D$. An evaluation~$\nu$ {\em satisfies} a constraint
$\langle (x_1,\ldots,x_n),R \rangle$ if $(v(x_1),\ldots,v(x_n)) \in R$. A
{\em solution} is an evaluation that satisfies all constraints.

\st
Consider an alphabet consisting of regular and constraint atoms,
denoted by $\ma$ and $\mc$ respectively. By
$\tilde{\mc}$, we denote the set of all literals over $\mc$. 
The constraint literals are identified with constraints via a function
$\gamma: \tilde{\mc} \rightarrow C$ so that for any literal
$l$, $\gamma(l)$ has a solution if and only if $\gamma(\overline l)$ does not have
one (where $\overline l$ denotes a complement of $l$). 
For a set $Y$ of constraint literals over $\mc$,
by $\gamma(Y)$ we denote a set of corresponding constraints, i.e., 
$\{\gamma(c) \,\,|\,\,  c\in Y\}$.  Furthermore, each variable in
$\gamma(Y) $  is associated with a domain. 
For a set $M$ of literals, by~$M^+$ and ~$M^{\mc}$  we denote  the set of
positive  literals in $M$ and  the set of
constraint literals over $\mc$ in $M$, respectively.

\st
A {\em logic program with constraint atoms} is a regular logic program
over an extended alphabet $\ma\cup \mc$ such that, in rules of the form
(\ref{e:rule}),
 $a_0$ is $\bot$ or $a_0\in\ma$. Given a logic program with
constraint atoms~$\Pi$, by $\Pi^{\mc}$ we denote $\Pi$ extended
with choice rules $\{c\}$ for each constraint atom $c$ occurring in
  $\Pi$.
We say that  a consistent and complete set $M$  of literals over atoms
of $\Pi$ is an {\em answer set} of $\Pi$ 
 if  
\begin{enumerate}
\labitem{(a1)}{as.1}  $M^+$ is an answer set of $\Pi^{\mc}$ and
\labitem{(a2)}{as.2}  $M^{\mc}$ has a solution.
\end{enumerate}
The CASP problem is the problem of
determining, given  a logic program with
constraint atoms~$\Pi$, whether  $\Pi$ has an answer set.

\st
For example, let $\Pi$ be the program 
\begin{equation}\label{ex:acp}
\ba l
  am\ar\ X< 12 \\
  lightOn\ar\ switch, \no am\\
  \{switch\}\\
 \bot  \ar \no lightOn. 
\ea
\end{equation}
Intuitively, this program states that (a) {\it light} is {\it on}
 only if an action of {\it switch}
occurs during the {\it pm} hours and 
(b) {\it light} is {\it on} (according to the last
rule in the program). Consider a domain of $X$ to be integers from $0$ till
$24$. It is easy to see that a set
$$ 
\{switch, lightOn, \neg\ am, \neg X<12 \}
$$
%\textbf{TODO: $\neg$ is not introduced above}
forms the only answer set of program~(\ref{ex:acp}). 

\st
One may now draw a parallel to satisfiability modulo theories
(SMT). To do so we first formally define the SMT problem. 
A {\em theory} $T$ is a set of closed first-order
formulas. A CNF formula~$F$ (a set of clauses) is {\em $T$-satisfiable} if $F\wedge T$ is satisfiable in the
first-order sense. 
Otherwise, it is called $T$-unsatisfiable.
Let $M$ be a set of literals. We sometimes may identify $M$ with a
conjunction consisting of all its elements. We say that  $M$ is a
$T$-model of~$F$ if 
\begin{enumerate}
\labitem{(m1)}{m.1} $M$ is a model of $F$ and 
\labitem{(m2)}{m.2} 
$M$ is  {\em  $T$-satisfiable}. 
\end{enumerate}
The SMT problem for a theory $T$ is the problem of
determining, given a formula $F$, whether~$F$ has a $T$-model.
It is easy to see that in the CASP problem, $\Pi^{\mc}$ in condition~\ref{as.1}  plays the role of~$F$ in~\ref{m.1}  in the SMT
problem. At the same time, the
condition~\ref{as.2} is similar in nature to the condition~\ref{m.2}.

\st
Given this tight conceptual relation between the SMT and CASP formalisms,
it is not surprising that  solvers stemming from these different
research areas share a lot in common in their design even though these areas
have been developing to a large degree independently (CASP being a much
younger field). 
We start next section by reviewing major design principles/methods in crafting
 SMT solvers. We then discuss how  CASP solvers follow one or
 another method.
This discussion allows us to systematize solvers' design patterns
present both in SMT and CASP so that their relation
becomes clearer. Such transparent view on solvers' architectures 
immediately  translates  findings in one area into another. Thus
although  the case study conducted in this research uses CASP
technology only, we expect similar results to hold for
SMT, and for the construction of hybrid automated reasoning methods in
general.

\section{SMT/CASP Integration Schemas}\label{sec:integration}
Satisfiability modulo theories (SMT) integrates different theories 
``under one roof''. Often it also integrates different
computational procedures for processing such hybrid theories.
We are interested in these synergic procedures explored by the SMT
community over
the past decade. We follow~\cite[Section 3.2]{nie06} for a review of 
several integration techniques exploited in SMT. 
% and also borrow their
%terminology.  

\st
In every discussed approach, 
 a formula $F$ is treated as a satisfiability formula
where each of its atoms is considered as a
propositional symbol, {\it forgetting} about the theory $T$. Such view
naturally invites an idea of {\em lazy} integration: the formula
$F$ is given to a SAT solver, if the solver determines that $F$ is
unsatisfiable then $F$ is $T$-unsatisfiable as well. Otherwise, a
propositional model $M$ of $F$ found by the SAT solver is checked by a
specialized $T$-solver which determines whether~$M$ is 
$T$-satisfiable. If so, then it is also a $T$-model of~$F$,
otherwise~$M$ 
is used to build a clause $C$ that precludes this assignment,
i.e., $M\not\models C$ while $F \cup C$ is $T$-satisfiable if and only
if $F$ is $T$-satisfiable. The SAT solver is invoked on an augmented
formula $F\cup C$. Such process is repeated until the procedure finds
a $T$-model or returns unsatisfiable.  Note how in this approach two
automated reasoning systems -- a SAT solver and a specialized
$T$-solver -- interleave: a SAT solver generates ``candidate
models'' whereas  a $T$-solver tests whether these models are in
accordance with requirements specified by theory $T$. We find that it
is convenient to introduce the following terminology for the future
discussion: a {\it base} 
solver and a {\it theory} solver, where  a base solver is responsible
for generating candidate models and {\it theory} solver  is responsible
for any additional testing required for stating whether a candidate
model is indeed a solution. 

\st
It is easy to see how the lazy integration policy
translates into the realm of CASP. Given a program with constraint
atoms~$\Pi$, an answer set solver serves the role
of a base solver by generating answer sets of $\Pi^\mc$ 
(that are ``candidate answer sets'' for~$\Pi$) and then uses a
CLP/CSP solver as a theory solver to verify whether
condition~\ref{as.2} is satisfied on these candidate answer sets. 
Constraint answer set solver {\ezcsp} embraces the lazy integration
approach in its design.\footnote{\cite{bal09a} refers to 
lazy integration of {\ezcsp} as {\em lightweight}
integration of ASP and
constraint programming.} 
To  solve the CASP problem, {\ezcsp} offers a
user several options for  {\it base}  and  {\it theory} solvers. For
instance, it allows for the use of answer set solvers {\clasp}~\cite{geb07}, {\cmodels}~\cite{giu06}, {\dlv}~\cite{cit97} as
base solvers and CLP systems
{\sc SICStus Prolog}~\cite{sicstus} and {\sc Bprolog}~\cite{zho12} as theory solvers. Such variety in possible
configurations of {\ezcsp} illustrates how lazy integration
provides great flexibility in choosing underlying {base} and {theory}
solving technology in addressing problems of interest.

\st
The Davis-Putnam-Logemann-Loveland (DPLL) procedure~\cite{dav62} is a
backtracking-based search algorithm for deciding the satisfiability of
a propositional CNF formula. DPLL-like procedures form the basis for
most modern SAT solvers as well as answer set solvers. If a DPLL-like
procedure underlies a base solver in the SMT and CASP tasks then it
opens a door to several refinements of lazy integration. We now
describe these refinements that will also be a focus of the present case
study.

\st
In the lazy integration approach a base solver is
invoked iteratively. Consider the SMT task: 
a CNF formula $F_{i+1}$ of the $i+1^{\hbox{th}}$ iteration to a SAT solver
consists of a CNF formula $F_{i}$ of the $i^{\hbox{th}}$ iteration and an
additional clause (or a set of clauses). 
Modern DPLL-like solvers commonly implement such
technique as {\em incremental} solving. For instance,
incremental SAT-solving allows the user to solve several SAT problems
$F_1,\dots,F_n$ one after another (using single invocation
of the solver), if $F_{i+1}$
results from $F_i$ by adding clauses. In turn, 
the solution to~$F_{i+1}$ may benefit 
from the knowledge obtained
during solving  $F_1,\dots,F_i$.
Various modern SAT-solvers, including {\sc minisat} \cite{minisat,minisat-manual},
  implement interfaces for incremental 
SAT solving.
% and {\sc lingeling}http://fmv.jku.at/lingeling/. 
Similarly, the answer set solver {\cmodels} implements an interface
that  allows the user to solve several ASP problems
$\Pi_1,\dots,\Pi_n$ one after another, if $\Pi_{i+1}$
results from $\Pi_i$ by adding a set of rules whose heads are $\bot$. 
It is natural to utilize
incremental {\sc dpll}-like procedures for enhancing the lazy integration
protocol: we call this refinement
{\em lazy+} integration. In this approach rather than invoking a base
solver from scratch an incremental interface provided by a solver is
used to implement the iterative process.

\st
\cite{nie06} also reviews such integration techniques
used in SMT
as {\it on-line SAT solver} and {\it theory propagation}.
In the on-line SAT solver approach, the $T$-satisfiability of the (partial)
assignment is checked incrementally, while the assignment is being
built by the DPLL-like procedure. This can be done fully eagerly as
soon as a change in the partial assignment occurs or at some regular
intervals, for instance. Once the inconsistency is detected, a SAT
solver is instructed to backtrack.  The theory propagation
approach extends the on-line SAT solver technique by allowing
a theory solver  not only to verify that a current
partial assignment is $T$-consistent but also to detect literals in
a CNF formula that must hold given the current partial assignment.  
The CASP solver  {\clingcon} exemplifies the implementation of 
the theory propagation 
integration schema in CASP by unifying answer set solver {\sc clasp}
as a base solver and constraint processing system {\sc
  gecode}. %~\cite{gecode}. 
 \acsolver and \idp systems are other CASP solvers that implement 
the theory propagation integration schema.

\smallskip
\noindent
%\st
{\bf Three Kinds of \ezcsp:}
To conduct our analysis of various integration schemas and their
effect on the performance of the hybrid systems we used the CASP solver {\ezcsp}
as a baseline technology. As mentioned earlier, original  {\ezcsp}
implements the lazy integration schema. In the course of this work we 
developed enhanced interfaces with answer set solver {\cmodels} that
allowed for the two other integration schemas: lazy+ integration and on-line SAT solver.
These implementations
rely on API interfaces provided by
{\cmodels} 
%(using {\minisat} v. 1.12b) 
allowing for varying level of integration
between the solvers. The development of these API interfaces in
{\cmodels} was greatly facilitated by the API interface provided by
{\minisat} v. 1.12b supporting non-clausal constraints~\cite{minisat-manual}.
In the following we call
\begin{itemize}
\item {\ezcsp}
implementing lazy integration with {\cmodels} as a base solver -- a \bbox.
\item {\ezcsp}
implementing lazy+ integration with {\cmodels} --  a \gbox.
\item {\ezcsp}
implementing on-line  SAT solver integration with {\cmodels} (fully eagerly) -- a \cbox.
\end{itemize}
In all these configurations of \ezcsp we assume {\bprolog} to serve in the role
of a theory solver.

\section{Application Domains}\label{sec:domain}
In this work we compare and contrast different integration schemas of
hybrid solvers  on three application domains
that stem from various subareas of computer science. 
This section provides a
brief overview of these applications. 
All benchmark domains are 
from the {\sl Third  Answer Set Programming Competition --
  2011} ({\aspcomp})~\cite{aspcomp11}, 
in particular, 
 the {\sl Model and Solve} track. 
%Within this track the teams were 
%free to choose a specific
%declarative solver and modeling technique for each problem. 
We chose these domains for our investigation as they display
features that benefit from the synergy of computational methods in ASP
and CSP. Each considered problem contains variables ranging over a
large integer domain thus making grounding required in pure ASP a
bottleneck. On the other hand, the modeling capabilities
of ASP and availability of such sophisticated solving technique as learning
makes ASP attractive for designing solutions to these domains. 
As a result,  CASP languages and  solvers become a natural choice for
these benchmarks making them ideal for our investigation. 

\smallskip
\noindent
%\st
{\bf Three Kinds of CASP Encodings:} It is easy to note that hybrid languages
such as CASP  allow for mix-and-match
constructs and processing techniques stemming from different
formalisms. Yet, any pure ASP encoding of a problem is also a CASP
formalization of the same problem. Similarly, it is possible
to encode a problem in such a way that only the CSP solving
capabilities of the CASP paradigm are employed. In this study for
two of the benchmarks we considered  three kinds of encodings in the 
CASP language of {\ezcsp}: {\em pure-ASP} encoding; {\em pure-CSP} encoding; and
{\em true-CASP} encoding. In the third benchmark, the use of three distinct encodings was not possible because
both ASP and CSP features play a major role in the efficiency of the computation.

\st
Analysis of these varying kinds of encodings in CASP gives us a better
perspective on how different integration schemas are effected by the
design choices made during the encoding of a problem. At the same time
considering the encoding variety allows us to verify our
intuition that true-CASP is an appropriate modeling and solving choice
for the explored domains.    

\smallskip
\noindent
%\st
The {\bf weighted-sequence} (\wseq) domain is a handcrafted benchmark 
problem.
Its key features are inspired by the important 
industrial problem of finding an optimal join order by cost-based query
optimizers in database systems. 
\cite{lierPadl12} provides a
 complete description of the problem itself as well as the formalization
 that became ``golden standard'' in this work, i.e., the
 formalization named \seq. 

\st
In the weighted-sequence problem we are given a set of leaves (nodes) and an
integer $m$ -- maximum cost. Each leaf is a pair {\em (weight, cardinality)} where
{\em weight} and {\em cardinality} are integers. Every sequence (permutation) of leaves is
such that all leaves but the first are assigned a {\em color} that, in
turn, associates a leaf with a {\em cost} (via a cost formula). 
A colored sequence is
associated with the {\em cost} that is a sum of leaves' costs. 
The task is to find a colored sequence
with cost at most $m$. We refer the reader to \cite{lierPadl12}
for the details of pure-ASP encoding \seq. The same paper also
contains the details on a true-CASP variant of \seq in the language of
{\clingcon}. We further adapted that encoding to the language of {\ezcsp} by
means of simple syntactic transformations. Here we provide a 
review of details of the \seq formalization that we find most relevant to this presentation.
The non-domain predicates of the pure-ASP encoding are 
$\mathit{leafPos}$, $posColor$, $posCost$. Intuitively,
 $\mathit{leafPos}$ is responsible for assigning a position to a
leaf,  $posColor$ is responsible for assigning a color to each position,
 $posCost$ carries information on costs associated with
each leaf.
The main difference between the
 pure-ASP and true-CASP encodings is in the treatment of the cost values of
the leaves. We first note that  cost  predicate
 $posCost$  in the pure-ASP encoding
 is "functional". In other words, 
when this predicate occurs in an answer
set its first argument uniquely determines its second argument.
 Often, such functional
predicates in ASP encodings can be replaced by constraint atoms in
CASP encodings. Indeed, this is the case in the weighted-sequence
problem. Thus in the true-CASP encoding, predicate $posCost$
is replaced by constraint atoms, making it possible to evaluate
cost values by CSP techniques. This approach is expected to
benefit performance especially when the cost values are large.
Predicates $\mathit{leafPos}$ and $posColor$
are also functional.
The pure-CSP encoding is obtained from the true-CASP encoding by
replacing $\mathit{leafPos}$ and $posColor$ predicates by constraint
atoms.

%In that work, the authors investigated the
% behavior of pure ASP solver {\clingo} versus CASP solver
% {\clingcon}. Their findings support the claim that for instances
% where weight variable ranges over large integer domain CASP
% technology provides computational advantage. 
%Neverthless, from the
% conducted experimental analysis  it is
% evident that exploiting learning available in ASP is crucial.

\smallskip
\noindent
%\st
The {\bf incremental scheduling} (\is) domain stems from a problem
occurring in commercial
printing. In this domain,  a schedule is maintained up-to-date with respect to 
 jobs being added and  equipment going offline.
A problem description includes a set of
devices, each with predefined number of instances (slots for jobs), and a set of 
jobs to be produced. The penalty for a 
job being tardy is computed as $td \cdot imp$, where $td$ 
is the job's tardiness and $imp$ is a positive integer denoting 
the job's importance.
The total penalty of a schedule is the sum of the penalties 
of the jobs.
The task is to find a schedule whose
total penalty is no larger than the value  specified in
 a problem instance.
We direct the reader to~\cite{bal11} for a complete
description of the domain.
The pure-CSP encoding used in our experiments is the
official competition encoding submitted
to {\aspcomp} by the {\ezcsp} team.
In this encoding, constraint atoms are used for (i) 
assigning start times to jobs, (ii) selecting which
device instance will perform a job, and (iii) 
calculating tardiness and penalties.
The true-CASP encoding was obtained from the pure-CSP encoding
by introducing a new relation $on\_instance(j,i)$, stating
that job $j$  runs on device-instance $i$. 
This relation and ASP
constructs of the \ezcsp language replaced the constraint 
atoms responsible for the assignment of device instances in the
pure-CSP encoding.
The pure-ASP encoding was obtained from the true-CASP encoding
by introducing suitable new relations, such as $start(j,s)$
and $penalty(j,p)$, to replace all the remaining constraint atoms.

%For example, selecting a large press sheet would prevent the use of a small press. The
%underlying decision-making process is often called production planning (the term
%"planning" here is only loosely related to the meaning of planning the execution
%of actions over time typical in the ASP community, but is retained because it is
%relatively well established in the field of the application). Another set of decisions
%deals with scheduling. Here one needs to determine when the various jobs will

\smallskip
\noindent
%\st
In the {\bf reverse folding} (\rf) domain,
one manipulates a sequence of $n$
pairwise connected segments located on a 2D plane in order to take the sequence from
an initial configuration to a goal configuration.
The sequence is manipulated by pivot moves: rotations of a segment around its
starting point by 90 degree in either direction.
A pivot move on a segment causes
the segments that follow to rotate around the same
center. 
Concurrent pivot moves are prohibited. At the end of each
move, the segments in the sequence must not intersect.
A problem instance specifies the number of segments, 
the goal configuration, and required number of moves, $t$.
The task is to find a sequence of exactly
$t$ pivot moves that produces the goal configuration.
The true-CASP encoding used for our experiments is from
the official {\aspcomp2011} submission package of the {\ezcsp} team.
In this encoding, relation $pivot(s,i,d)$ states that at step
$s$ the $i^{th}$ segment is rotated in direction $d$.
The effects of pivot moves are described by constraint
atoms, which allow carrying out the corresponding calculations
with CSP techniques.
The pure-ASP encoding was obtained from the true-CASP encoding
by adopting an ASP-based formalization of the effects of
pivot moves. This was accomplished by introducing two new
relations, $\mathit{tfoldx}(s,i,x)$ and $\mathit{tfoldy}(s,i,y)$,
stating that the new start of segment $i$ at step $s$
is $\tbeg x, y \tend$.
The definition of the relations is provided by
suitable ASP rules.

\section{Experimental Results}\label{sec:experiments}
%\begin{figure}[t]
%\begin{center}
%\includegraphics[scale=0.33]{CHART-LINKS/performance-chart-wseq-true_CASP.eps}\ %%\ \ 
%\\
%\ \\
%\includegraphics[scale=0.33]{CHART-LINKS/performance-chart-wseq-totals.eps}
%\end{center}
%\caption{Performance on {\wseq} domain: (a) true-CASP encoding; (b) total %times in logarithmic scale}
%\label{fig:wseq}
%%\label{fig:wseq-true-casp}
%%\label{fig:wseq-totals}
%\end{figure}
\begin{figure}[t]
\begin{center}
\includegraphics[scale=0.50]{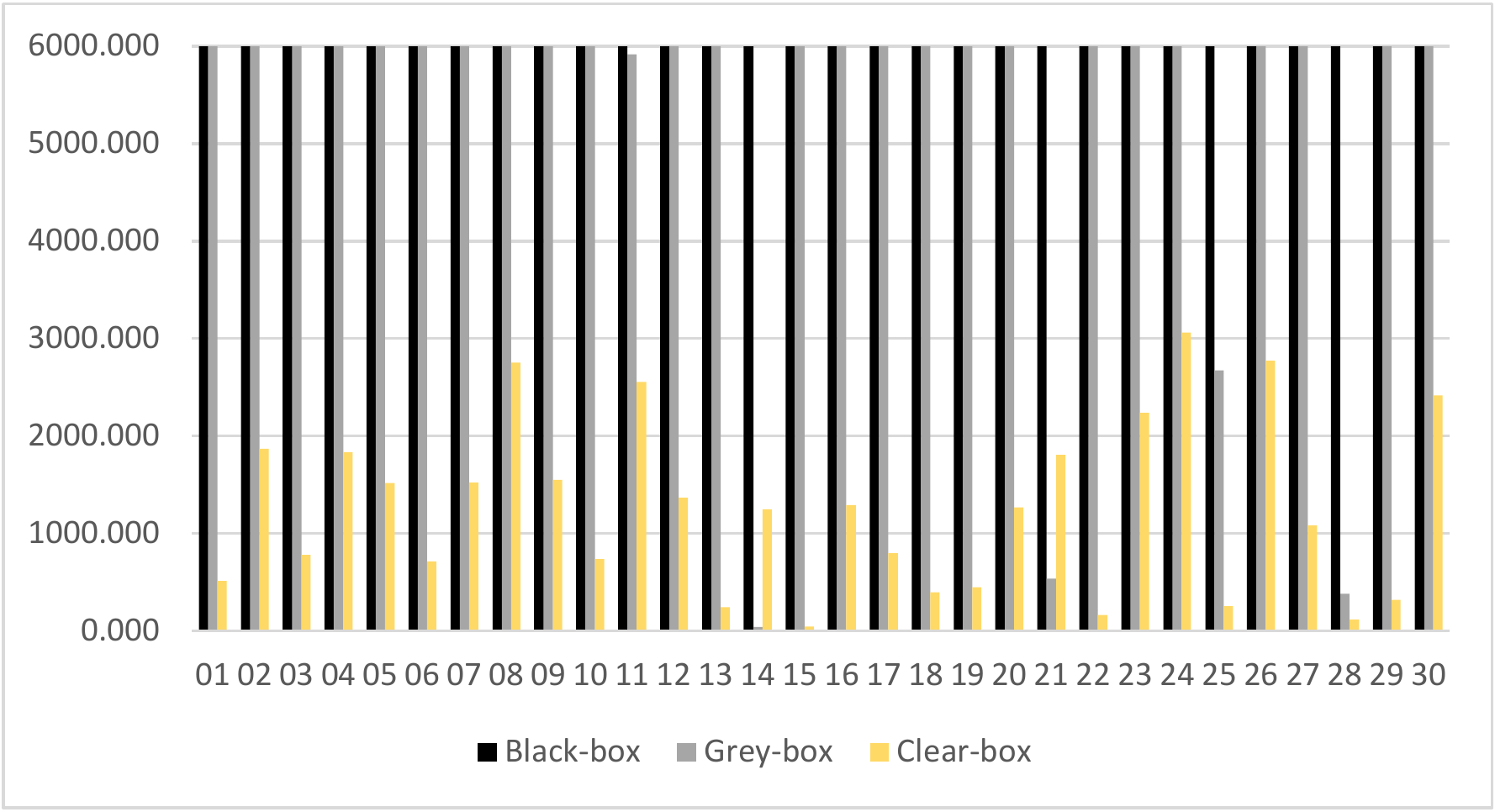}\ \end{center}
\caption{Performance on {\wseq} domain: true-CASP encoding}
%\label{fig:wseq}
\label{fig:wseq-true-casp}
\end{figure}
\begin{figure}[t]
\begin{center}
\includegraphics[scale=0.50]{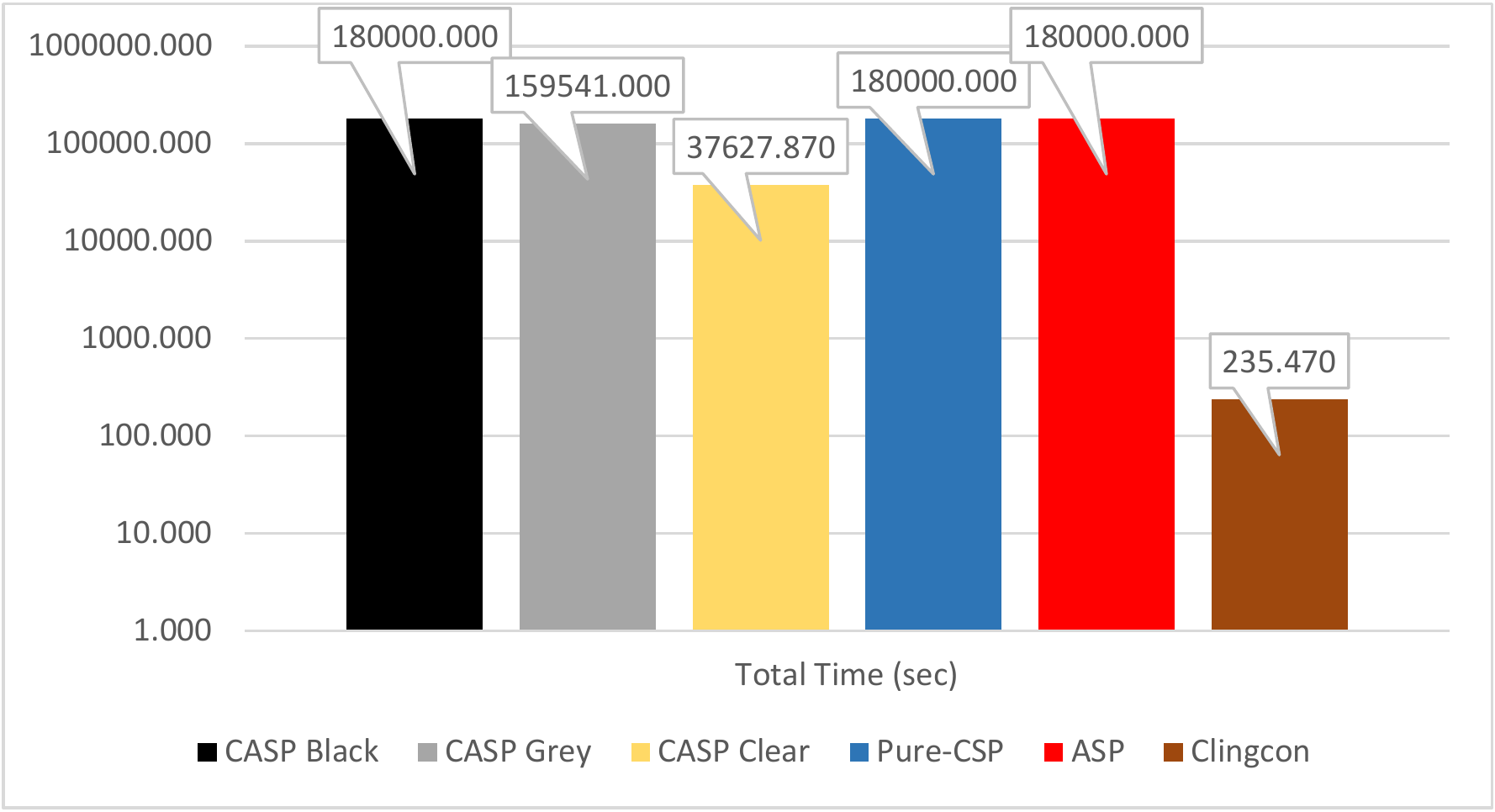}
\end{center}
\caption{Performance on {\wseq} domain: total times in logarithmic scale}
%\label{fig:wseq}
\label{fig:wseq-totals}
\end{figure}

\st
The experimental comparison of the integration schemas
was conducted on a computer with an Intel Core i7
processor running at 3GHz. Memory limit for each process 
and timeout considered were 
 $3$ GB
RAM and $6,000$ seconds respectively.
The version of {\ezcsp}
used in the experiments was 1.6.20b49: it incorporated {\cmodels} version
3.83 as a base solver
and  {\bprolog} 7.4\_3 as a theory solver. 
Answer set solver {\cmodels} 3.83 was also used for the
experiments with
 the pure-ASP encodings.
%The results are summarized in the tables presented in the Appendix.
In order to provide a frame of reference with respect
to the state of the art in CASP, the tables for \wseq and
\is include performance information for {\clingcon} 2.0.3
on  true-CASP encodings adapted to the language of {\clingcon}.
The ezcsp executable used in the experiments and the
encodings can be downloaded from 
\emph{http://www.mbalduccini.tk/ezcsp/aspocp2013/ezcsp-binaries.tgz}
and
\emph{http://www.mbalduccini.tk/ezcsp/aspocp2013/experiments.tgz}
respectively.
In all figures presented:
\begin{itemize}
\item  CASP Black, CASP Grey, CASP Clear denote 
{\ezcsp} implementing respectively \bbox, \gbox and \cbox, and
running a true-CASP encoding; 
\item Pure-CSP denotes {\ezcsp} implementing \bbox running
a pure-CSP encoding (note that for pure-CSP encodings there is no
difference in performance between the integration schemas); 
\item ASP denotes {\cmodels} running a pure-ASP encoding;
\item  Clingcon  denotes
{\clingcon} running a true-CASP encoding.
\end{itemize}
%\textbf{EXPLAIN WHY WE DIDN'T RUN CLINGCON ON \rf.}\\
%\textbf{WHERE DO WE SAY THAT ASP WAS SLOW EVERYWHERE?}

\st
We begin our analysis with \wseq.
The instances used in the experiments are the 30
instances available via {\aspcomp}.
{\wseq} proves to be a domain that truly
requires the interaction of the ASP and CSP solvers.
Answer set solver {\cmodels} on the pure-ASP encoding
runs out of memory on every instance (in the tables, 
out-of-memory conditions and timeouts are
both rendered as out-of-time results).
\ezcsp on the pure-CSP encoding  
reaches the timeout limit on every instance.
The true-CASP encoding running in \bbox also times out on every instance. 
\st 
As shown in 
%Figure~\ref{fig:wseq}(a),
Figure~\ref{fig:wseq-true-casp},
the true-CASP encoding running in \gbox performs slightly
better. %, since the search space of the ASP solver is re-used
%when an answer set is found that violates the CSP constraints.
The true-CASP encoding running in \cbox instead performs \emph{substantially}
better. 
%We believe that this is due to the fact that the tighter integration 
%of the ASP and CSP solvers makes it possible to prune the search space of
%the ASP solver while the computation of an answer set is still
%on-going.
%Figure~\ref{fig:wseq}(b) 
Figure~\ref{fig:wseq-totals} 
reports the total times across all
the instances for all solvers/encodings pairs considered. Notably,
CASP solver {\clingcon} on true-CASP encoding is several orders of
magnitude faster than any other configuration. 
This confirms that
for this domain tight integration schemas indeed have
an advantage. Recall that {\clingcon} implements a tighter integration
schema than that of {\ezcsp} {\cbox} that, in addition to the on-line SAT
  solver schema of \cbox, also includes theory
    propagation.  Answer set solver {\clasp} serves the role of base
    solver of {\clingcon} whereas {\gecode} is the theory solver. 
%We believe that these differences explain the experimental results.

%\begin{figure}[h]
%\begin{center}
%\includegraphics[scale=0.33]{CHART-LINKS/performance-chart-is-easy-totals.eps}
%%\ \ 
%\\
%\ \\
%\includegraphics[scale=0.245]{CHART-LINKS/performance-chart-is-easy-global.eps}
%%\ \ \\
%\\
%\ \\
%\includegraphics[scale=0.33]{CHART-LINKS/performance-chart-is-easy-true_CASP.eps}
%\end{center}
%\caption{Performance on {\is} domain, easy instances: (a) total times (ASP encoding off-chart); (b) overall view; (c) true-CASP encoding (detail of 0-1sec execution time range)}
%\label{fig:is-easy}
%\label{fig:is-easy-totals}
%\label{fig:is-easy-overall}
%\label{fig:is-easy-true_CASP}
%\end{figure}
\begin{figure}[h]
\begin{center}
\includegraphics[scale=0.50]{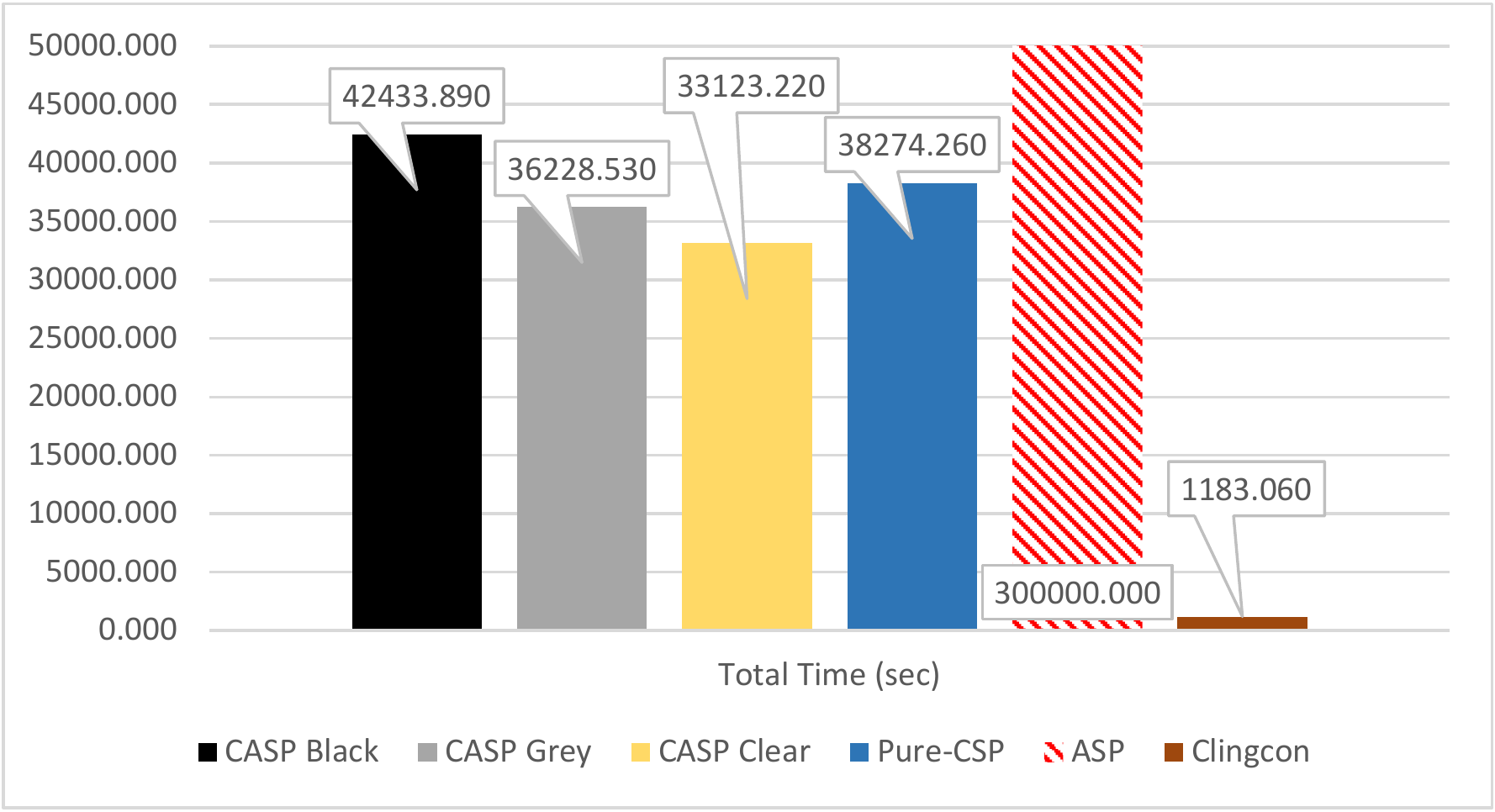}
\end{center}
\caption{Performance on {\is} domain, easy instances: total times (ASP encoding off-chart)}
%\label{fig:is-easy}
\label{fig:is-easy-totals}
\end{figure}
\begin{figure}[h]
\begin{center}
\includegraphics[scale=0.37]{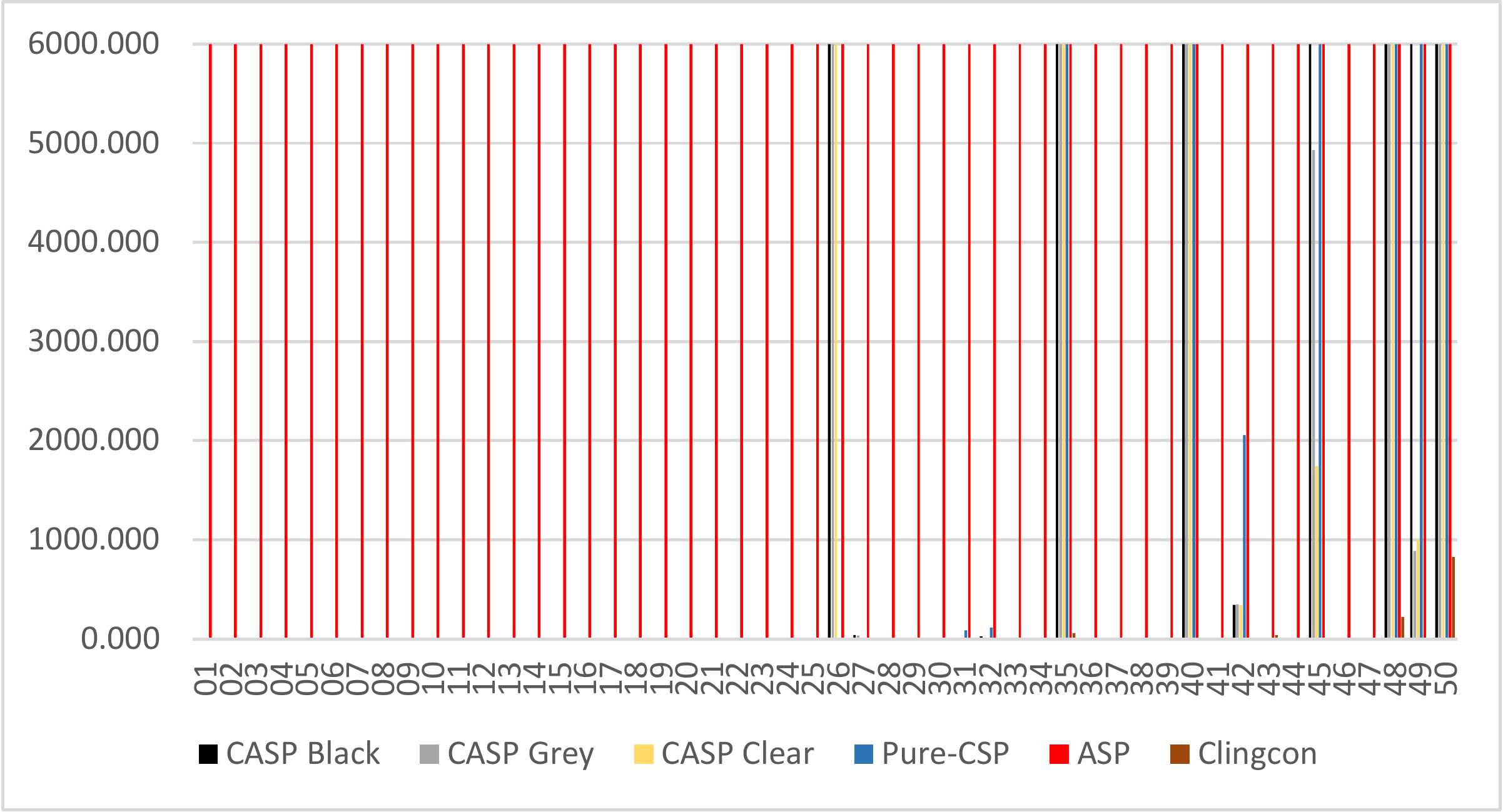}
\end{center}
\caption{Performance on {\is} domain, easy instances: overall view}
%\label{fig:is-easy}
\label{fig:is-easy-overall}
\end{figure}
\begin{figure}[h]
\begin{center}
\includegraphics[scale=0.50]{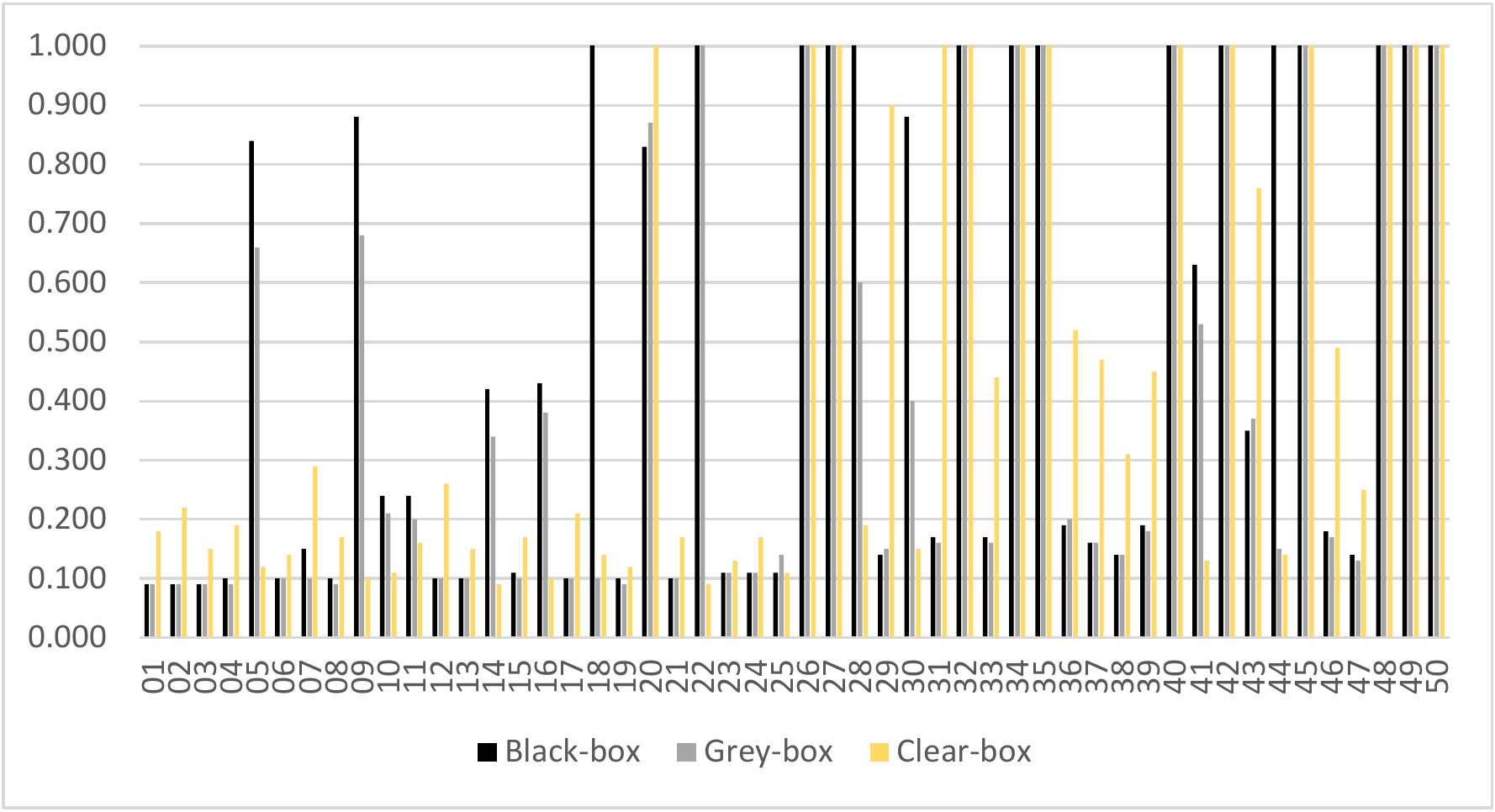}
\end{center}
\caption{Performance on {\is} domain, easy instances: true-CASP encoding (detail of 0-1sec execution time range)}
%\label{fig:is-easy}
\label{fig:is-easy-true_CASP}
\end{figure}

%\begin{figure}[h]
%\begin{center}
%\includegraphics[scale=0.33]{CHART-LINKS/performance-chart-is-hard-totals.eps}
%%\ \ 
%\\
%\ \\
%\includegraphics[scale=0.33]{CHART-LINKS/performance-chart-is-hard-global.eps}
%%\ \ \\
%%\includegraphics[scale=0.33]{CHART-LINKS/performance-chart-is-hard-true_CASP.eps}
%\end{center}
%\caption{Performance on {\is} domain, hard instances: (a) overall view;
%(b) total times
%%; (c) true-CASP encoding
%}
%\label{fig:is-hard}
%%\label{fig:is-hard-totals}
%%\label{fig:is-hard-overall}
%\end{figure}
\begin{figure}[h]
\begin{center}
\includegraphics[scale=0.50]{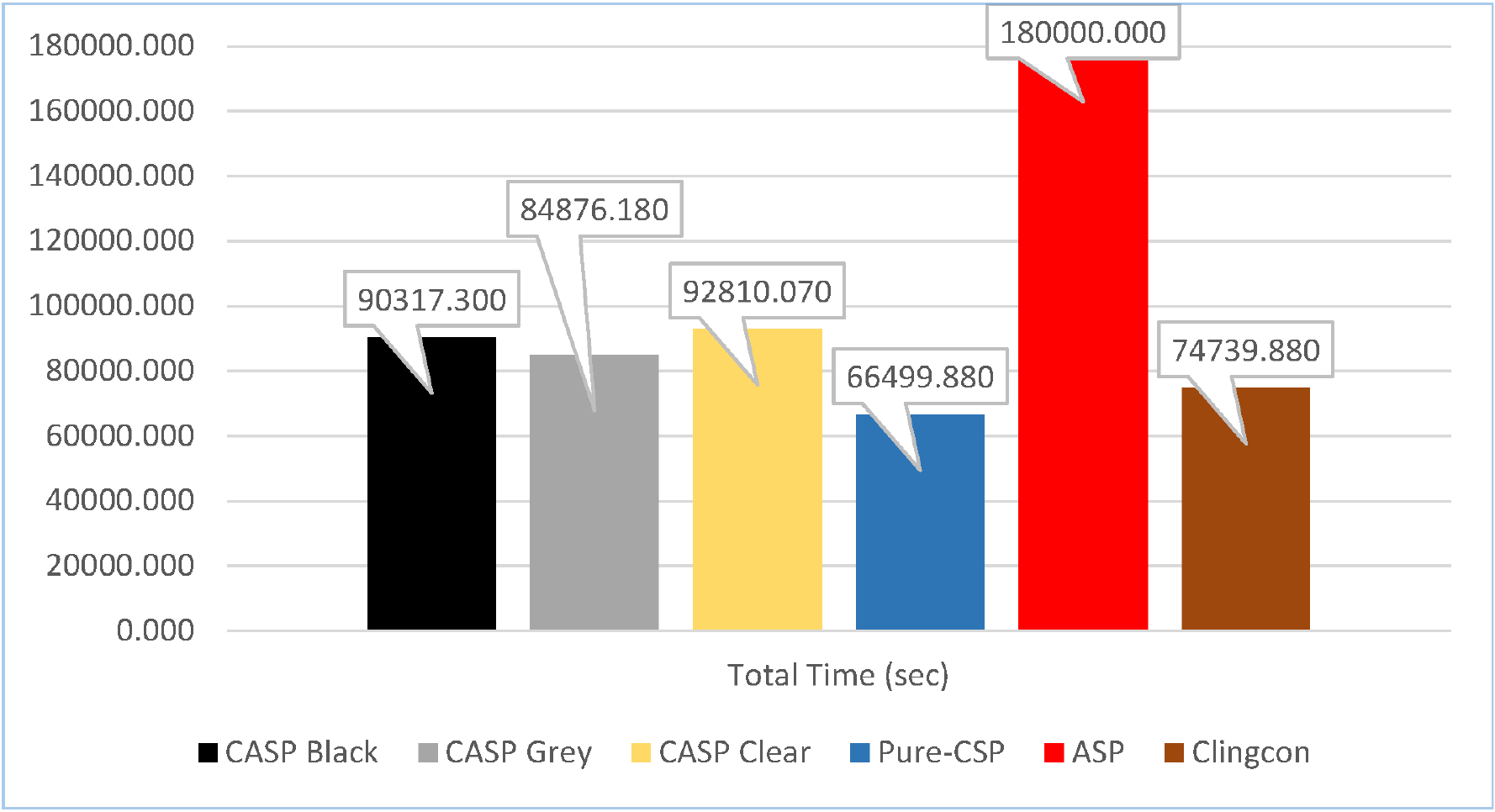}
\end{center}
\caption{Performance on {\is} domain, hard instances: overall view}
%\label{fig:is-hard}
\label{fig:is-hard-totals}
\end{figure}
\begin{figure}[h]
\begin{center}
\includegraphics[scale=0.50]{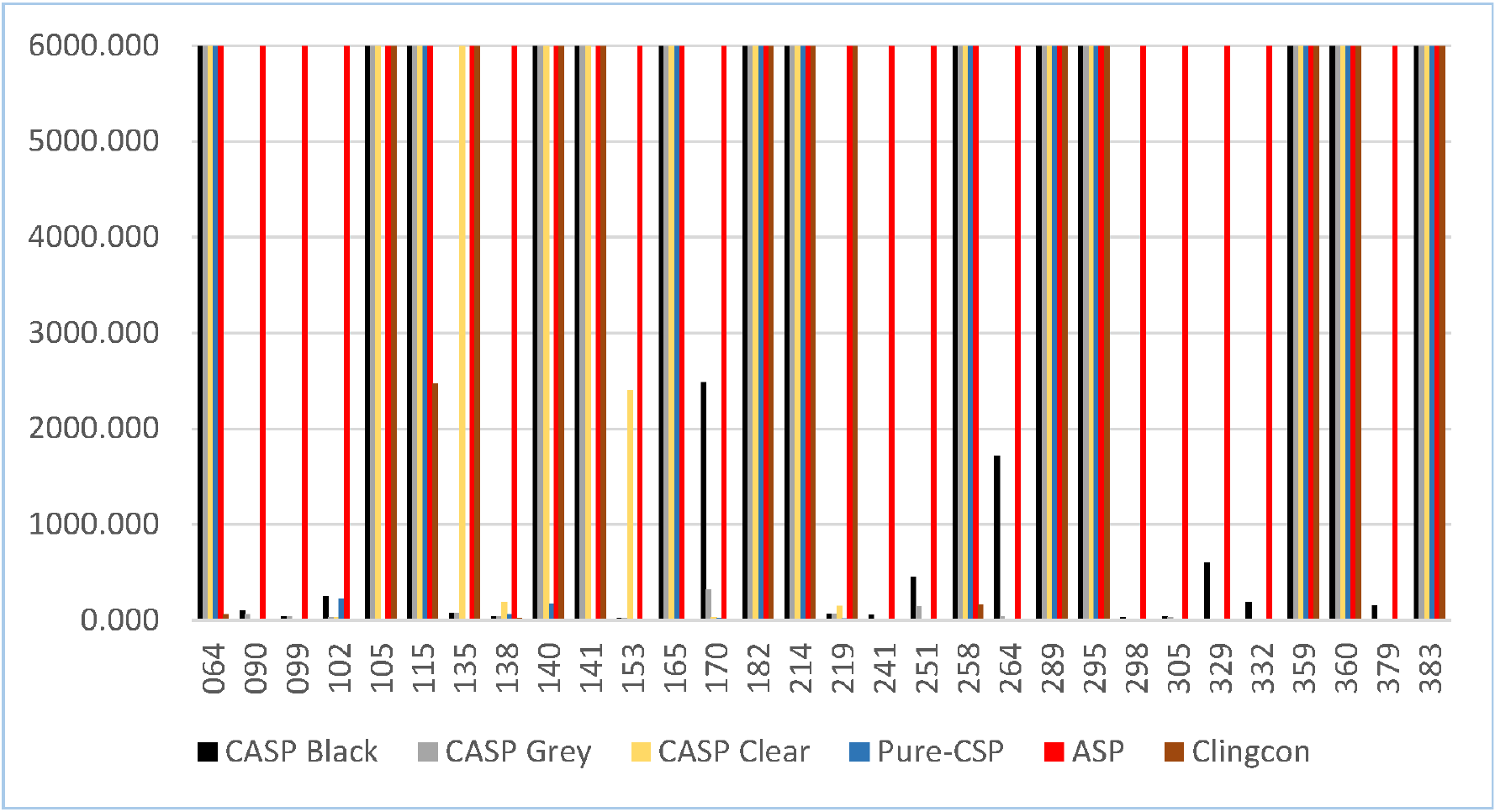}
\end{center}
\caption{Performance on {\is} domain, hard instances: total times}
%\label{fig:is-hard}
\label{fig:is-hard-overall}
\end{figure}
%\begin{figure}[h]
%\begin{center}
%\includegraphics[scale=0.75]{CHART-LINKS/performance-chart-is-hard-true_CASP.eps}
%\end{center}
%\caption{Performance on {\is} domain, hard instances; true-CASP encoding}
%\label{fig:is-hard-true_CASP}
%\end{figure}
%\begin{figure}[h]
%\begin{center}
%\includegraphics[scale=0.75]{CHART-LINKS/performance-chart-is-hard-pure_CSP.eps}
%\end{center}
%\caption{Performance on {\is} domain, hard instances; pure-CSP encoding}
%\label{fig:is-hard-pure_CSP}
%\end{figure}

\st
In case of the {\is} domain we considered two sets of experiments. In
the former we
 used the 50 official instances from {\aspcomp}. We refer to these
 instances as {\em easy}.
%Figure~\ref{fig:is-easy-overall} given in Appendix 
Figure~\ref{fig:is-easy-overall} 
%Figure~\ref{fig:is-easy}(b) 
depicts the overall per-instance performance
on the {\is}-easy domain.
It appears that tight integration schemas
have an advantage, allowing the true-CASP encoding to outperform
the pure-CSP encoding. As one might expect, the
best performance for the true-CASP encoding is obtained
with the {\cbox} integration schema, as shown in 
%Figure~\ref{fig:is-easy-totals} 
%as well as in Figure~\ref{fig:is-easy-true_CASP} presented in Appendix.
Figure~\ref{fig:is-easy-totals} 
and in Figure~\ref{fig:is-easy-true_CASP}.
%Figure~\ref{fig:is-easy}(a) 
%and Figure~\ref{fig:is-easy}(c).
Figure~\ref{fig:is-easy-totals} 
%Figure~\ref{fig:is-easy}(a) 
provides a comparison of the total times.
In this case the early pruning of the search space 
made possible by the {\cbox} architecture yields substantial benefits.
As expected, it is also the case that {\gbox} is faster than {\bbox}.
As for {\wseq}, \clingcon is the fastest, and {\cmodels} on
the pure-ASP encoding runs out of memory in all the instances.

\st
The next experiment reveals an interesting change in behavior of
solver/encodings pairs
as the complexity of the instances of the  {\is} domain grows.
In this test, we used a set of $30$ instances
obtained by (1) generating randomly $500$ fresh instances;
(2) running the true-CASP
encoding with the {\gbox} integration schema on them
with a timeout of $300$ seconds; (3) 
selecting randomly, from those, $15$ instances that resulted
in timeout and $15$ instances that were solved in $25$ seconds
or more. 
The numerical parameters used in the process
were selected with the purpose of identifying challenging instances.
The overall per-instance execution times reported in
Figure~\ref{fig:is-hard-overall} 
%Figure~\ref{fig:is-hard}(b) 
clearly indicate the
level of difficulty of the selected instances.
Remarkably, these more difficult instances
are solved more efficiently by the pure-CSP encoding that relies
only on the CSP solver, as evidenced by  
Figure~\ref{fig:is-hard-totals}.
%Figure~\ref{fig:is-hard}(a).
In fact, the pure-CSP encoding outperforms every other method of computation,
\emph{including} {\clingcon} on true-CASP encoding.
More specifically, solving the instances with the true-CASP 
encoding takes between $30\%$ and
$50\%$ longer than with the pure-CSP encoding.
(Once again, {\cmodels} runs out of memory.)
%In line with this result, the {\cbox} integration
%schema is the slowest among the schemas running the
%pure-CSP encoding, because of the overhead due to
%the more frequent interactions by the ASP and the CSP solver.

%\begin{figure}[h]
%\begin{center}
%\includegraphics[scale=0.33]{CHART-LINKS/performance-chart-rf-totals.eps}
%%\ \ 
%\\
%\ \\
%\includegraphics[scale=0.2438]{CHART-LINKS/performance-chart-rf-global.eps}
%%\ \ \\
%\\
%\ \\
%\includegraphics[scale=0.33]{CHART-LINKS/performance-chart-rf-true_CASP.eps}
%\end{center}
%\caption{Performance on {\rf} domain: (a) total times (detail of 0-1000sec execution time range, ASP and {\cbox} off-chart); (b) overall view;
%(c) true-CASP encoding, detail of 0-0.50sec execution time range}
%\label{fig:rf}
%%\label{fig:rf-totals}
%%\label{fig:rf-overall}
%%\label{fig:rf-true_CASP}
%\end{figure}
\begin{figure}[h]
\begin{center}
\includegraphics[scale=0.50]{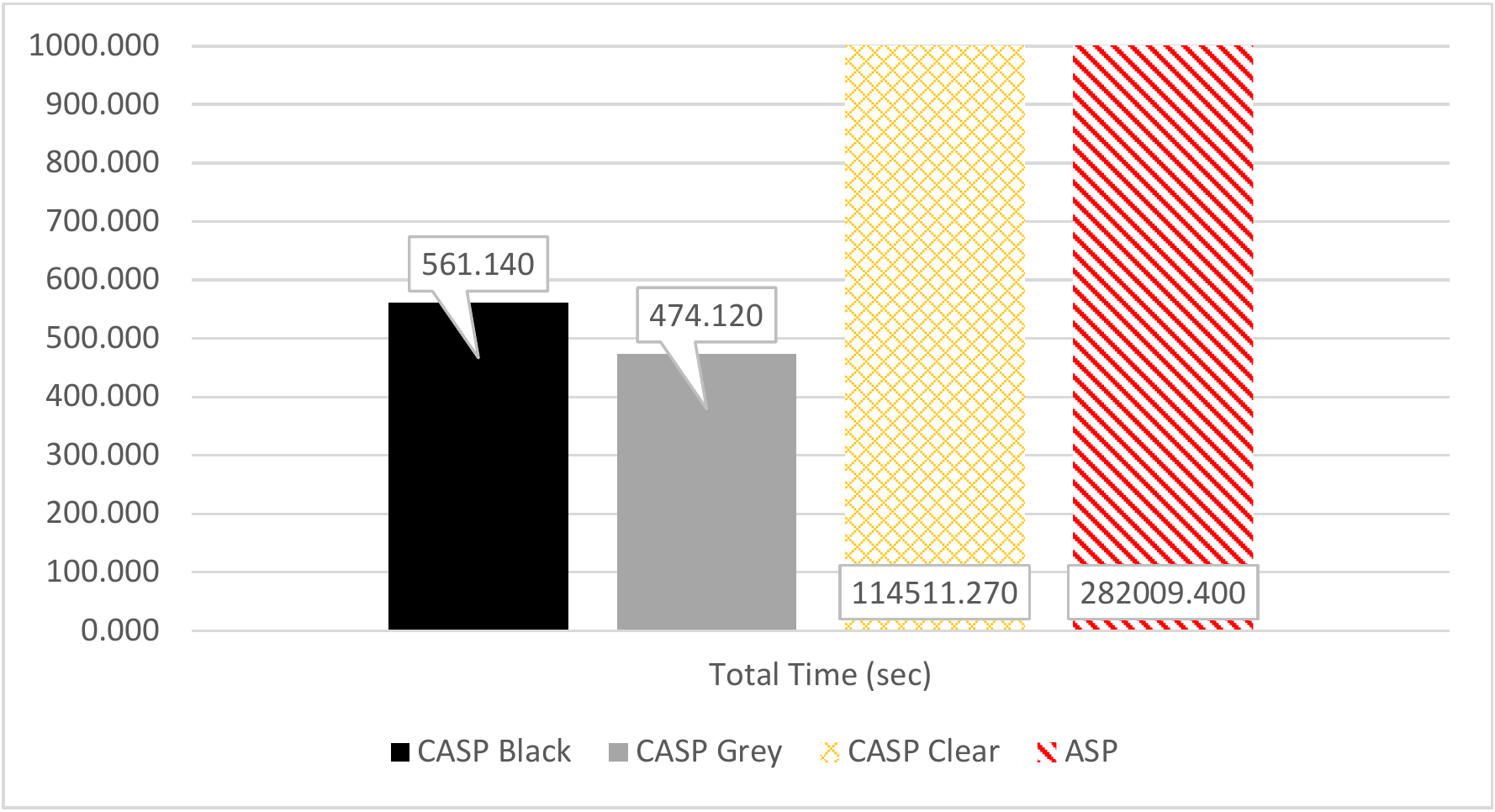}
\end{center}
\caption{Performance on {\rf} domain: total times (detail of 0-1000sec execution time range, ASP and {\cbox} off-chart)}
%\label{fig:rf}
\label{fig:rf-totals}
\end{figure}
\begin{figure}[h]
\begin{center}
\includegraphics[scale=0.37]{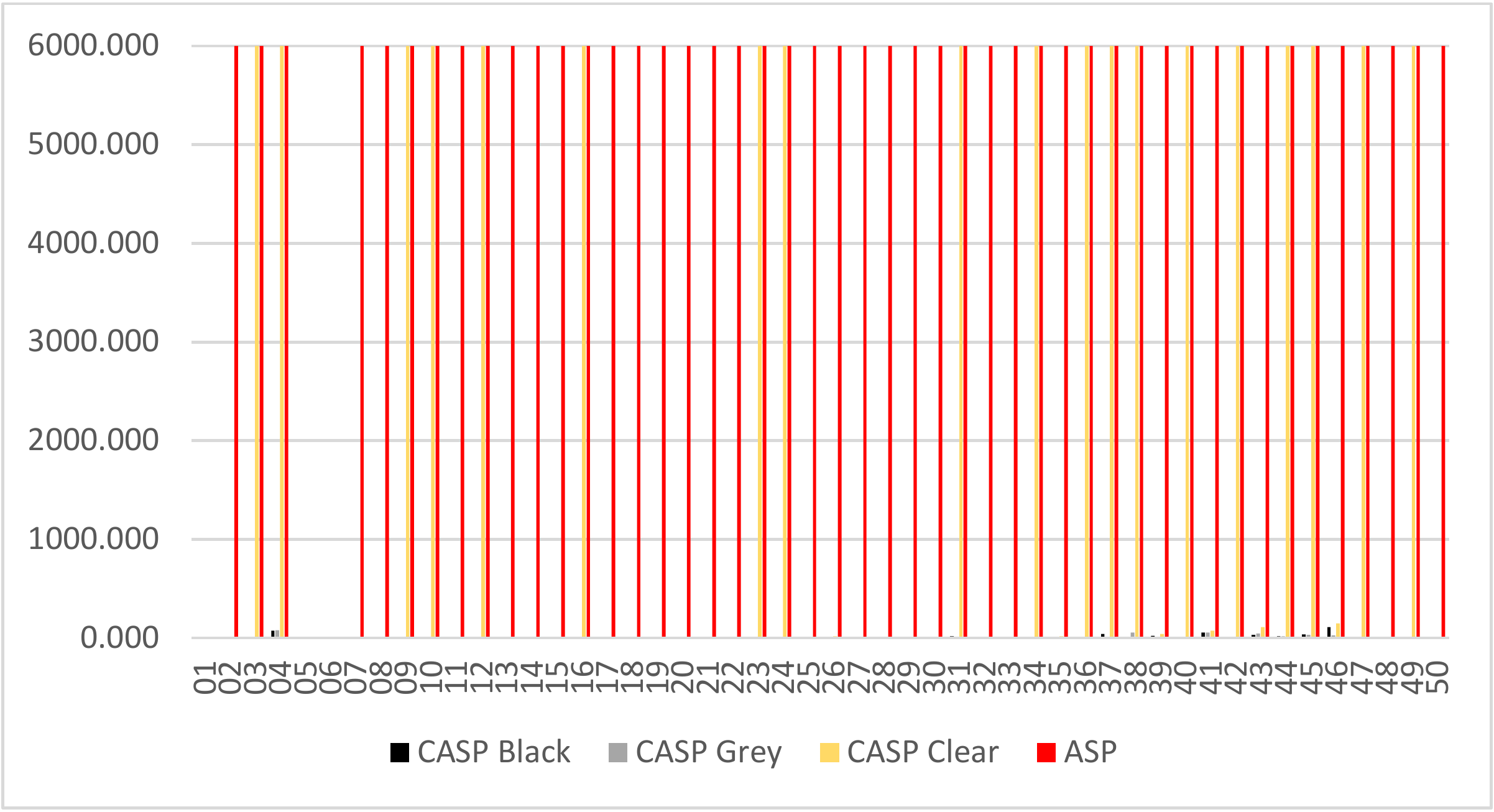}
\end{center}
\caption{Performance on {\rf} domain: overall view}
\label{fig:rf-overall}
\end{figure}
\begin{figure}[h]
\begin{center}
\includegraphics[scale=0.50]{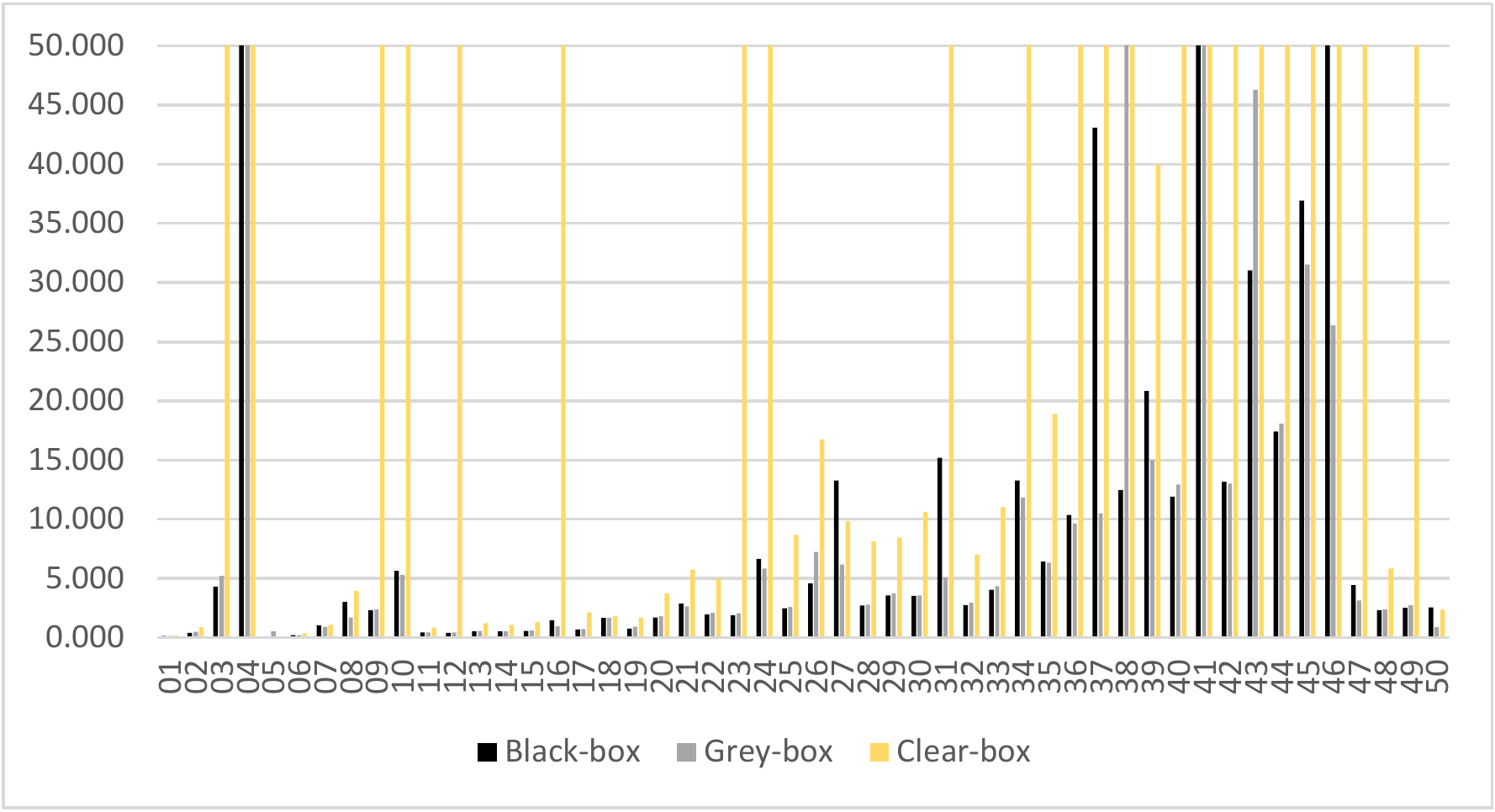}
\end{center}
\caption{Performance on {\rf} domain: true-CASP encoding, detail of 0-0.50sec execution time range}
%\label{fig:rf}
\label{fig:rf-true_CASP}
\end{figure}

\st
The final experiment focuses on the {\rf} domain.  We
 used the 50 official instances from {\aspcomp} to conduct the analysis.
%Figure~\ref{fig:rf-overall} presented in the Appendix 
Figure~\ref{fig:rf-overall} presented 
%Figure~\ref{fig:rf}(b)
shows that this domain is
comparatively easy. 
Figure~\ref{fig:rf-true_CASP} 
%Figure~\ref{fig:rf}(c) 
illustrates that 
the {\bbox} and {\gbox}
integration schemas are
several orders of magnitude faster than {\cbox}. 
This somewhat surprising result can be explained by the fact that 
in this domain frequent checks
with the theory solver add more overhead rather than being of help in
identifying an earlier point to backtrack.
{\cmodels} on the pure-ASP encoding runs out of memory in all but $3$
instances.
The total execution times are presented in 
Figure~\ref{fig:rf-totals}.
%Figure~\ref{fig:rf}(a).

\section{Conclusions}
The case study conducted in this work clearly illustrates the
influence that integration methods have on the behavior of hybrid 
systems. Each integration schema may be of use and importance for some
domain. Thus systematic means ought to be found for facilitating
building   hybrid  systems supporting various coupling
mechanisms. Building clear and flexible API interfaces allowing for
various types of interactions between the solvers seems a necessary 
step towards making the development of hybrid solving
systems effective.
This work provides evidence for the need of an effort to 
this ultimate goal.

%\newpage
\bibliographystyle{splncs03}
\bibliography{biblio,biblio-yulia-fixed}

\begin{thebibliography}{10}
\providecommand{\url}[1]{\texttt{#1}}
\providecommand{\urlprefix}{URL }

\bibitem{aspcomp11}
Third answer set programming competition (2011),
  \url{https://www.mat.unical.it/aspcomp2011/}

\bibitem{bal09a}
Balduccini, M.: Representing constraint satisfaction problems in answer set
  programming. In: Proceedings of ICLP'09 Workshop on Answer Set Programming
  and Other Computing Paradigms (ASPOCP'09) (2009)

\bibitem{bal11}
Balduccini, M.: {I}ndustrial-{S}ize {S}cheduling with {A}{S}{P}+{C}{P}. In:
  Delgrande, J.P., Faber, W. (eds.) 11th International Conference on Logic
  Programming and Nonmonotonic Reasoning (LPNMR11). Lecture Notes in Artificial
  Intelligence (LNCS), vol. 6645, pp. 284--296. Springer Verlag, Berlin (2011)

\bibitem{bre11}
Brewka, G., Niemel\"{a}, I., Truszczy\'{n}ski, M.: Answer set programming at a
  glance. Communications of the ACM  54(12),  92--103 (2011)

\bibitem{cit97}
Citrigno, S., Eiter, T., Faber, W., Gottlob, G., Koch, C., Leone, N., Mateis,
  C., Pfeifer, G., Scarcello, F.: The {\sc dlv} system: Model generator and
  application frontends. In: Proceedings of Workshop on Logic Programming
  (WLP97) (1997)

\bibitem{dav62}
Davis, M., Logemann, G., Loveland, D.: A machine program for theorem proving.
  Communications of the ACM  5(7),  394--397 (1962)

\bibitem{minisat}
Een, N., Biere, A.: Effective preprocessing in sat through variable and clause
  elimination. In: SAT (2005)

\bibitem{minisat-manual}
Een, N., S\"{o}rensson, N.: An extensible sat-solver. In: SAT (2003)

\bibitem{fer05b}
Ferraris, P., Lifschitz, V.: Weight constraints as nested expressions. Theory
  and Practice of Logic Programming  5,  45--74 (2005)

\bibitem{geb07}
Gebser, M., Kaufmann, B., Neumann, A., Schaub, T.: Conflict-driven answer set
  solving. In: Proceedings of 20th International Joint Conference on Artificial
  Intelligence (IJCAI'07). pp. 386--392. MIT Press (2007)

\bibitem{geb09}
Gebser, M., Ostrowski, M., Schaub, T.: Constraint answer set solving. In:
  Proceedings of 25th International Conference on Logic Programming (ICLP). pp.
  235--249. Springer (2009)

\bibitem{giu06}
Giunchiglia, E., Lierler, Y., Maratea, M.: Answer set programming based on
  propositional satisfiability. Journal of Automated Reasoning  36,  345--377
  (2006)

\bibitem{gom08}
Gomes, C.P., Kautz, H., Sabharwal, A., Selman, B.: Satisfiability solvers. In:
  van Harmelen, F., Lifschitz, V., Porter, B. (eds.) Handbook of Knowledge
  Representation, pp. 89--134. Elsevier (2008)

\bibitem{jm94}
Jaffar, J., Maher, M.: Constraint logic programming: A survey. Journal of Logic
  Programming  19(20),  503--581 (1994)

\bibitem{lier-alp}
Lierler, Y.: Constraint answer set programming  (2012),
  \url{http://www.cs.utexas.edu/users/ai-lab/pub-view.php?PubID=127221}

\bibitem{lier12}
Lierler, Y.: On the relation of constraint answer set programming languages and
  algorithms. In: Proceedings of the 26th Conference on Artificial Intelligence
  (AAAI-12). MIT Press (2012)

\bibitem{lierPadl12}
Lierler, Y., Smith, S., Truszczynski, M., Westlund, A.: Weighted-sequence
  problem: Asp vs casp and declarative vs problem oriented solving. In:
  Fourteenth International Symposium on Practical Aspects of Declarative
  Languages (2012),
  \url{http://www.cs.utexas.edu/users/ai-lab/pub-view.php?PubID=127085}

\bibitem{lif99d}
Lifschitz, V., Tang, L.R., Turner, H.: Nested expressions in logic programs.
  Annals of Mathematics and Artificial Intelligence  25,  369--389 (1999)

\bibitem{mel08}
Mellarkod, V.S., Gelfond, M., Zhang, Y.: Integrating answer set programming and
  constraint logic programming. Annals of Mathematics and Artificial
  Intelligence  (2008)

\bibitem{nie00}
Niemel{\"a}, I., Simons, P.: Extending the {Smodels} system with cardinality
  and weight constraints. In: Minker, J. (ed.) Logic-Based Artificial
  Intelligence, pp. 491--521. Kluwer (2000)

\bibitem{nie06}
Nieuwenhuis, R., Oliveras, A., Tinelli, C.: Solving {S}{A}{T} and {S}{A}{T}
  modulo theories: From an abstract {D}avis-{P}utnam-{L}ogemann-{L}oveland
  procedure to {D}{P}{L}{L}({T}). Journal of the ACM  53(6),  937--977 (2006)

\bibitem{ros08}
Rossi, F., van Beck, P., Walsh, T.: Constraint porgramming. In: van Harmelen,
  F., Lifschitz, V., Porter, B. (eds.) Handbook of Knowledge Representation,
  pp. 181--212. Elsevier (2008)

\bibitem{spe13}
Schuller, P., Patoglu, V., Erdem, E.: {A} {S}ystematic {A}nalysis of {L}evels
  of {I}ntegration between {L}ow-{L}evel {R}easoning and {T}ask {P}lanning. In:
  Workshop on Combining Task and Motion Planning at the IEEE International
  Conference on Robotics and Automation 2013 (2013)

\bibitem{sicstus}
SICStus: {S}icstus {P}rolog {W}eb {S}ite (2008),
  http://www.sics.se/isl/sicstuswww/site/

\bibitem{idp}
Wittocx, J., Mari\"en, M., Denecker, M.: The {\sc idp} system: a model
  expansion system for an extension of classical logic. In: LaSh. pp. 153--165
  (2008)

\bibitem{zho12}
Zhou, N.F.: {T}he language features and architecture of {B}-{P}rolog. Journal
  of Theory and Practice of Logic Programming (TPLP)  12(1--2),  189--218 (Jan
  2012)

\end{thebibliography}

\end{document}